\documentclass[11pt, a4paper, logo, copyright]{deepmind}

\pdfoutput=1

\usepackage[authoryear, sort&compress, round]{natbib}

\usepackage{microtype}
\usepackage{graphicx}
\usepackage{caption,subcaption}
\usepackage{hyperref}
\usepackage{wrapfig}

\renewcommand{\today}{2022-04-21}

\usepackage{amsmath}
\usepackage{amssymb}
\usepackage{mathtools}
\usepackage{amsthm}

\newcommand{\set}[1]{\ensuremath{\mathcal{#1}}}
\renewcommand{\S}{\set{S}}
\newcommand{\A}{\set{A}}

\newcommand{\messagedown}[1]{\Omega_{#1}}

\title{Learning how to Interact with a Complex Interface using Hierarchical Reinforcement Learning}

\correspondingauthor{gcomanici@deepmind.com glamia@deepmind.com kenjitoyama@deepmind.com}

\author[1,*]{Gheorghe Comanici }
\author[1,*]{Amelia Glaese}
\author[1,*]{Anita Gergely}
\author[1,*]{Daniel Toyama}
\author[1]{Zafarali Ahmed}
\author[1]{Tyler Jackson}
\author[1]{Philippe Hamel}
\author[1]{Doina Precup}

\affil[1]{DeepMind}
\affil[*]{Equal contribution}

\begin{abstract}
Hierarchical Reinforcement Learning (HRL) allows interactive agents to decompose complex problems into a hierarchy of sub-tasks. Higher-level tasks can invoke the solutions of lower-level tasks as if they were primitive actions.  In this work, we study the utility of hierarchical decompositions for learning an appropriate way to interact with a complex interface. Specifically, we train HRL agents that can interface with applications in a simulated Android device. We introduce a Hierarchical Distributed Deep Reinforcement Learning architecture that learns (1) subtasks corresponding to simple finger gestures, and (2) how to combine these gestures to solve several Android tasks. Our approach relies on goal conditioning and can be used more generally to convert any base RL agent into an HRL agent. We use the AndroidEnv environment to evaluate our approach. For the experiments, the HRL agent uses a distributed version of the popular DQN algorithm to train different components of the hierarchy. While the native action space is completely intractable for simple DQN agents, our architecture can be used to establish an effective way to interact with different tasks, significantly improving the performance of the same DQN agent over different levels of abstraction.
\end{abstract}


\begin{document}

\maketitle

\section{Introduction}
\label{sec:intro}

As we scale up Reinforcement Learning (RL) agents to tackle large varieties of problems in domains that are commonly controlled by humans, these agents need to consider how to acquire and reuse diverse knowledge about the world~\citep{ring1994continual, kakade2003sample, cisek2010neural, pezzulo2016navigating}.
AndroidEnv is an open-sourced domain that poses such a challenge: general purpose agents need to control a universal touchscreen interface and tackle a wide variety of tasks in Android applications; the latter are developed for human users, hence they leverage human abilities to reuse knowledge and and build intuitions through constant interaction with the platform~\citep{toyama2021androidenv}. Controlling AndroidEnv is purposely designed to match real devices: agents observe screen pixels and control finger positioning in real-time; the environment runs in its own timeline and does not wait for the agent to deliberate over its choices; actions are executed asynchronously; the agent has the potential to interact with any Android application. 

One of the main driving principles for Hierarchical Reinforcement Learning (HRL) is the explicit decomposition of RL problems into a hierarchy of subtasks such that higher-level parent-tasks invoke low-level child tasks as if they were primitive actions. The space of all possible decompositions is complex and hard to work with, albeit extensive research shows that proper inductive biases can be used to facilitate the search for useful decompositions (e.g. diffusion models~\citep{MachadoBB17}, bottleneck states~\citep{MenacheMS02, Simsek2004}, intrinsic goals~\citep{KulkarniNST16}, language~\citep{JiangGMF19}, empowerment~\citep{salge2014empowerment}). We introduce an HRL agent that acquires simple finger gesture skills and successfully reuses this knowledge in several diverse AndroidEnv tasks. To demonstrate the generality of the approach, we use the framework of General Value Functions (GVFs)~\citep{suttongvfs11} to capture domain knowledge about gestures for AndroidEnv. GVFs have been proposed in prior work as a way to capture diverse knowledge about the world in the form of long-term predictions associated with agent experience. GVFs can be learned incrementally using off-policy methods, and can be used to capture knowledge at different time-scales and levels of abstraction~\citep{whitethesis, Modayil14nexting, SuttonT04, SchaulR13}.

Our main contribution is a novel Hierarchical Distributed Deep Reinforcement Learning architecture for AndroidEnv. The architecture first builds a goal-conditioned deep model~\citep{SchaulHGS15} for GVFs that capture knowledge about simple finger gestures then it learns how to combine corresponding skills to solve several tasks from Android applications. Instead of using general RL agents to solve a complex problem directly, the architecture first decomposes it into a three-level hierarchy of sub-tasks: the lowest level (level 0) interacts with the screen to complete gestures (taps, swipes and flings), the next level provides the target gesture (e,g. where to tap, direction of a swipe), the final level decides which gesture amongst the three to execute to maximize per-step rewards. The same general RL agent is then used to solve decision making processes corresponding to each of the levels in the hierarchy. We demonstrate that even though the native action space is intractable for the baseline distributed DQN agent~\citep{mnih2015humanlevel}, the same agent becomes much more efficient when used to solve sub-tasks and to make abstract choices at higher levels in the hierarchy. 

\section{The architecture}
\label{sec:architecture}

AndroidEnv is an open-source platform for Reinforcement Learning (RL) research, hence it allows one to experiment with many of the applications in the Android ecosystem using reinforcement learning algorithms. The many algorithms that can potentially be employed are commonly studied using the mathematical formalism of Markov Decision Processes (MDPs) with state space $\S$, action space $\A$, and transition function $p: \S \times \A \to \mathcal{D}(\S)$.\footnote{We use the notation $\mathcal{D}(\cdot)$ for probability distributions over a set.} 
A task is usually specified using a reward function $r: \S \times \A \times \S \to \mathbb{R}$ and a discount value $\gamma \in [0,1]$, and the purpose of RL agents is to ``solve'' such tasks by finding policies $\pi: \S \to \A$ that maximize the discounted expected return $E_\pi [R_0 + \gamma R_1 + \gamma R_2 + \cdots \gamma^{t-1} R_t + \cdots ] $. The latter is usually denoted by $v_\pi$ and is known as the value function of a policy $\pi$. Similarly, the optimal value function is denoted by $v^* = \max_\pi v_\pi$.

\paragraph{General Value Functions (GVFs).}\cite{suttongvfs11} introduced a unified way to express long-term predictions for signals that are independent of task-specific rewards, under policies that are different from the agent's behavior, and under flexible state-dependent discounting schemes. GVFs are associated with tuples $\langle \gamma, C, \pi \rangle$, where $\gamma: \S \to [0,1]$ is known as a \emph{continuation function}, defined over all states $\S$ of an MDP, $C: \S \times \A \times \S \to \mathbb{R}$ is the cumulant function over MDP transitions, and $\pi: \S \to \mathcal{D}(\A)$ is a policy that generates an action distribution for each MDP state. The corresponding prediction is denoted by $v_{\pi, \gamma, C}$ and it is the expected cumulant-based return: 
$$ v_{\pi, \gamma, C}(s) = \mathbb{E} \left[  \sum_{t=0}^\infty \left( \prod_{i=1}^t \gamma(S_i) \right) C_i \; | \; S_0 = s, A_{0:\infty} \sim \pi \right].$$ 
We use $q_{\pi, \gamma, C}(s,a)$ for predictions that are conditioned both on the initial state $S_0=s$ and action $A_0=a$. 
Discounted expected returns area appealing because they all obey some form of a \emph{Bellman equation} which greatly facilitates estimation and are used to derive tractable objective functions for optimization based algorithms, such as gradient descent over deep networks. For simplicity, we describe below the Bellman equation for the optimal cumulant-based $q$-value:
\[ q^*_{\gamma, C}(s,a) = \sum_{s' \in S} p(s' |s,a) \left[ C(s,a,s') + \gamma(s) \max_{a'} q^*_{\gamma, C} (s', a') \right]. \]
\paragraph{Options.}The options framework is a popular formalism for temporally extended actions. A \emph{option} $\omega$ can start execution in any of the states in the initialization set $\mathcal{I}_\omega \subseteq \S$, and it used policy $\pi_\omega$ to select actions and $\beta_\omega : \S \to [0,1]$ to determine whether to terminate execution or not.~\cite{sutton1999between} demonstrate that using options along side actions turns an MDP problem into a Semi Markov Decision Process, which itself can be equipped with optimality value functions and equivalent Bellman equations, i.e. options can be interchangeably used as actions.

\begin{figure}[t]
    \centering
    \includegraphics[width=0.65\textwidth]{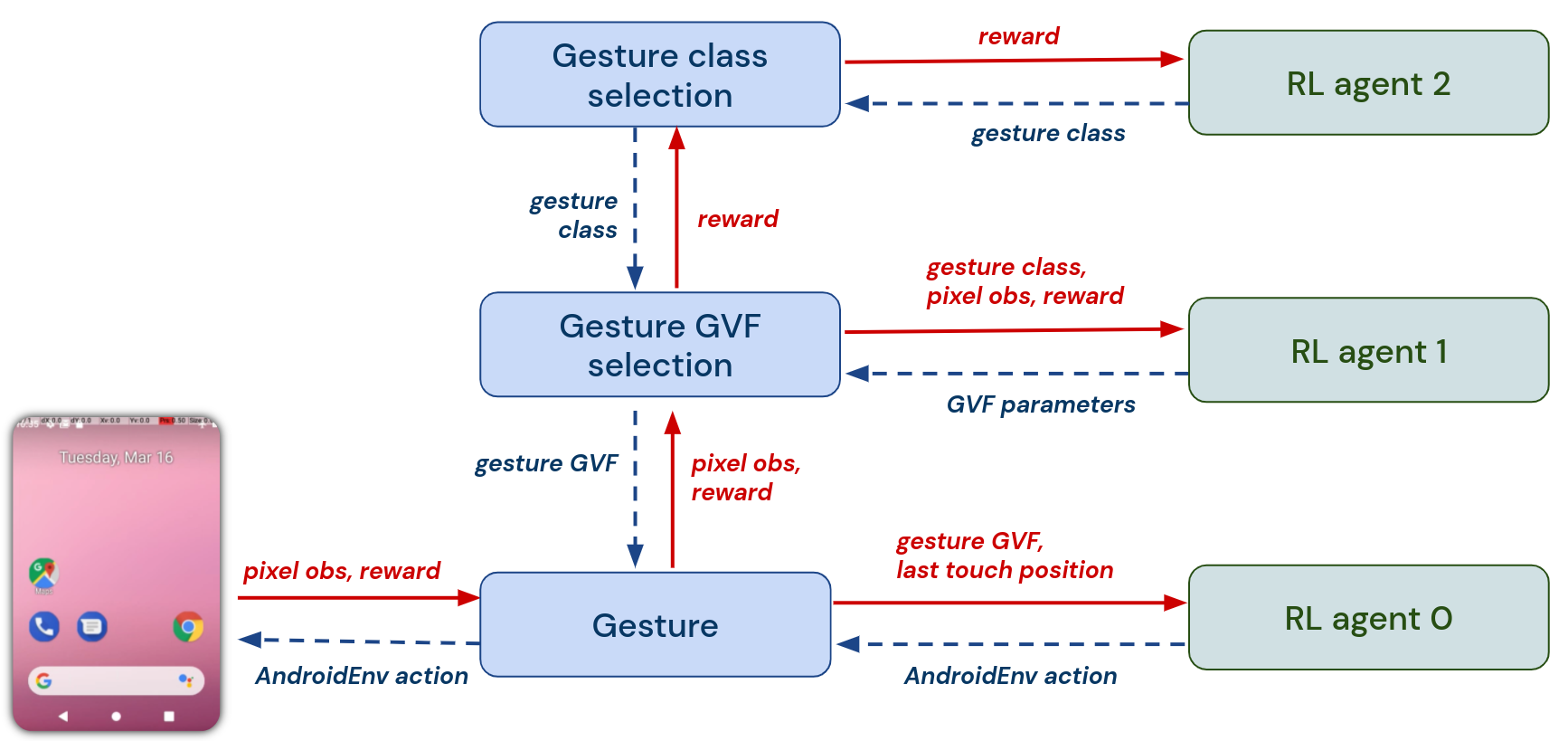}
    \caption{\textbf{Gesture Hierarchy}. The architecture used for the Android applications is based on a 3-layer hierarchy: (1) The lowest level operates over GVFs corresponding to all supported gestures; (2) The middle layer selects a gesture GVF given the latest pixel image in AndroidEnv and its agent is trained to maximize the return associated with the task that the agent is trained on; and (3) The top layer selects a single gesture class for the task and the agent is trained to maximize the average per step reward. All levels are operated by distributed DQN agents.}
    \label{fig:android:hierarchy}
\end{figure}

\paragraph{Hierarchy of GVFs.}
\label{sec:hierarchy}
We present a general approach to implement hierarchical decompositions of complex problems into a multi-layered hierarchy of sub-tasks, where each level is trained to maximize GVFs: given a fixed cumulant-continuation pair $(C,\gamma)$, agents maintain estimates for the value of the corresponding optimal policy, i.e. $q^*_{\gamma, C}(s,a) = \max_\pi q_{\pi, \gamma, C}(s,a)$. Instead of solving the problem with a single RL agent operating on the ``raw'' action space of an environment, we prioritize modularity and comprehension to build a hierarchy of ``problems'' that are solved by independent agents, working at different levels of space and temporal abstraction. 
A hierarchical decomposition on levels $0$ to $N$ works under the assumption that each level $i$ operates over a set of control GVFs,  $\messagedown{i} := \{ (C_i, \gamma_i) \}_{i=1}^{M}$ and, at each timestep, the corresponding RL agent follows the policy maximizing one of these GVFs. The selection of the active GVF at every timestep comes as a signal $\omega = (C, \gamma) \in \messagedown{i}$ from the level $i+1$. For all levels, except for the lowest level $0$, the corresponding agent selects an \emph{abstract} action $a_i$ by maximizing $q^*_{\gamma, C}(s, a_i)$, and propagates it down as a GVF selection for level $i-1$. In other words, the level is always maximizing one of the many signals that it is designed to predict. Lastly, temporal abstraction can be achieved within this framework by using the continuation function $\gamma$ of the selected GVF to determine the temporal extent of its execution. See Figure~\ref{fig:android:hierarchy} for the concrete three-level hierarchy we used in our work.
The main advantage of the hierarchical decomposition is that RL agents operating at different levels can be designed in isolation and perhaps can be trained either at different stages or using completely different techniques. For example, one could select among a finite set of abstract actions in level 1, while a continuous control agent interacts with an environment that operates with a continuous (or relatively large) action space. 

\begin{figure}
    \centering
    \includegraphics[width=0.27\textwidth]{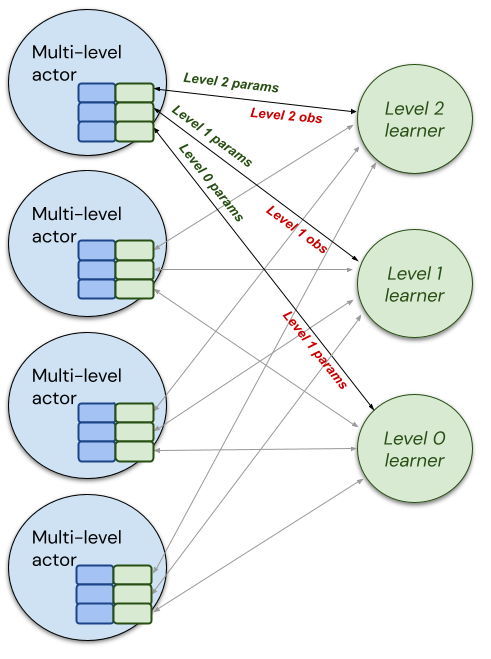}
    \caption{\textbf{Distributed hierarchies}. Multiple actors running on separate machines are used to generate data for $N$ different learners, one per level of hierarchy. For every interaction between a level $i$ and the corresponding RL agent interact, the observed interaction is communicated to the process maintaining the data for the Level $i$ learner. Periodically, actors retrieve the latest policy parameters from all learners.}
    \label{fig:hierarchy:distributed}
\end{figure}

\paragraph{Distributed Hierarchies.} Distributed computing architectures for Deep Reinforcement Learning have been shown to play an important role in scaling up these algorithms to relatively challenging domains~\citep{horgan18, KapturowskiOQMD19, OpenAI_dota, Jaderberg18}. In particular, these allow for asynchronous learning, and, when working with simulated environments, asynchronous acting. The modular hierarchical decomposition that we describe in this section is well suited for distributed architectures, as different levels operate with RL agents that are potentially independent of each other (see Figure \ref{fig:hierarchy:distributed}). Albeit these levels are tied during the execution of a policy due to the hierarchical signal processing procedure, learning is not: each level can maintain its own training dataset and perform learning updates on separate machines. Since AndroidEnv runs in real-time and the underlying simulation cannot be sped up, multiple actors run in parallel to generate sufficient experience for all learners.

\begin{figure}[t]
    \centering
    \includegraphics[width=0.8\textwidth]{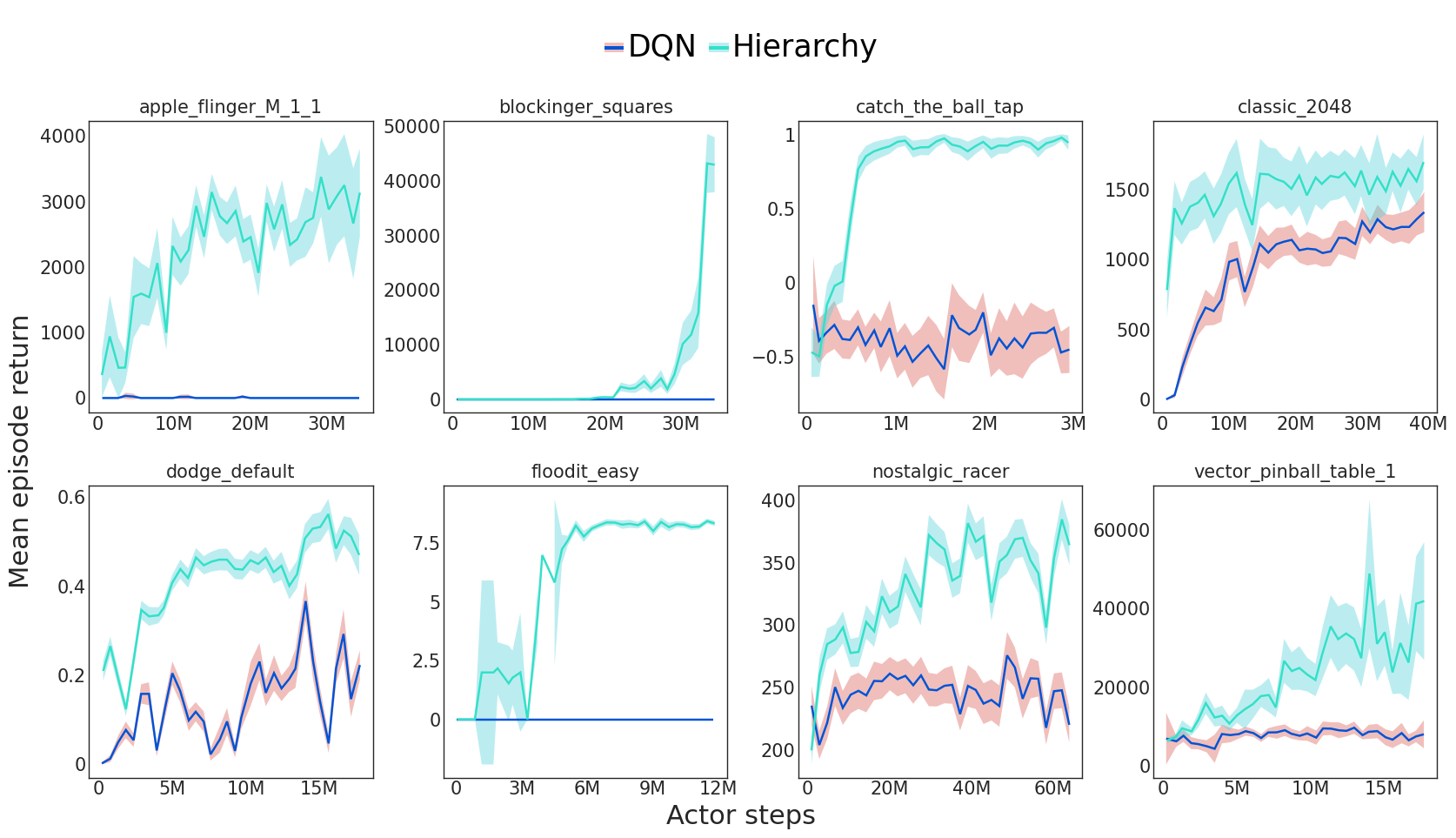}
    \caption{\textbf{Empirical results}. We tested our agents on a number of AndroidEnv tasks of different levels and with varying complexity in the action interface. We report results on tasks where at least one of the agents was able to improve its behavior. For tasks such as \texttt{classic\_2048} and \texttt{nostalgic\_racer}, using any fling or tap gesture, correspondingly, incurs significant changes in the score outcome. On the other hand, for tasks such as \texttt{apple\_flinger\_M\_1\_1}, \texttt{blockinger\_squares}, and \texttt{floodit\_easy}, the agent can only operate by direct interaction with specific buttons or objects and rewards are very sparse, making all of these tasks intractable for most agents.}
    \label{fig:empirical:apps}
\end{figure}

\section{Experimental implementation}
\label{sec:experiments}
We present results on a selection of AndroidEnv tasks. For our experiments, we used the Acme framework~\citep{hoffman2020acme} and its Distributed TensorFlow implementation of the DQN agent~\citep{mnih2015humanlevel}, configured for runs on Atari games, available at \href{https://github.com/deepmind/acme/tree/master/acme/agents/tf/dqn}{Acme's Github Repository}.\footnote{https://github.com/deepmind/acme/tree/master/acme/agents/tf/dqn} 
To be able to readily use agents designed for Atari games, we simplified the AndroidEnv interface by (1) down-sampling the input images to a 120 x 80 resolution, and (2) restricting taps and swipes to 54 locations on the screen, corresponding to a 9 by 6 discretization of the Android touch-screen. Moreover, the agent's input has further knowledge of any completed tap, swipe, or fling operation, as well as the most recent finger touch location. For more details on implementation, network architecture, and default hyper parameter settings, please refer to the Acme open-source code. Details on the set of AndroidEnv tasks for which we report results are available on \href{https://github.com/deepmind/android_env}{AndroidEnv's Github Repository.}\footnote{https://github.com/deepmind/android\_env} 
Figures~\ref{fig:empirical:apps} and~\ref{fig:empirical:apps:numbers} provide a summary of the observed empirical results. The rest of this section provides a detailed description of the hierarchy used to obtain these results.

\paragraph{Level 0: gesture execution.} 
The lowest level in the hierarchy is designed to execute gestures by operating on a set of GVFs composed of tap, swipe, and fling gestures. To fully define these GVFs, level 0 maintains a sequence of all touch positions in a trajectory, denoted by $( \textbf{p}_0, \textbf{p}_1 \cdots, \textbf{p}_t) $, with all $\textbf{p}_i$ either positions on the screen for tap actions or $\textbf{p}_i = \mathbf{0}$ for lift actions. For example, to capture a swipe gesture from location $\textbf{q}_1$ to $\textbf{q}_2$ we use a cumulant 
$$ C_{\textbf{q}_1, \textbf{q}_2}( \textbf{p}_0, \textbf{p}_1 \cdots, \textbf{p}_t) = \begin{cases} 1 & \mbox{ if } \exists i < t \mbox{ with } [ \textbf{p}_i, \textbf{p}_{i+1}, \dots, \textbf{p}_{t-1}, \textbf{p}_t] = [\mathbf{0}, \textbf{q}_1, \textbf{p}_{i+2}, \dots, \textbf{p}_{t-2}, \textbf{q}_2, \mathbf{0}] \\
 & \mbox{ and }\textbf{p}_{j} \neq \textbf{0}, \forall i< j < t, \\ 
0 & \mbox{ otherwise.} \end{cases} $$
The continuation function is set to $\gamma_{\textbf{q}_1, \textbf{q}_2} = 1 - C_{\textbf{q}_1, \textbf{q}_2}$. In all experiments, we use tap locations and swipe start/end locations based on the $9$ by $6$ discretization described above, resulting in $54 x 54$ swipe GVFs and $54$ tap GVFs. We additionally define $8$ fling GVFs corresponding to $N, NE, E, SE, S, SW, W$ and $NW$ cardinal directions.

As illustrated in Figure~\ref{fig:android:hierarchy}, the signal from above fully define individual gestures: $\omega_0 \in \Omega_0$ contains both a gesture class and a gesture parameter, e.g. $\omega_0 = (\mbox{swipe}, \textbf{q}_1, \textbf{q}_2)$ for a swipe from $\textbf{q}_1$ to $\textbf{q}_2$. To train the corresponding agent, we concatenate one-hot encodings for the gesture class, gesture parameters, and the last tap location. Each class of gestures was trained separately, hence the execution at this level is based on 3 separate networks. Lastly, we also apply Hindsight Experience Replay (HER)~\citep{Andrychowicz17} for improved data-efficiency: we always select a single GVF during acting, but we compute cumulants and continuations for all GVFs as to relabel the training data and use it to train predictions corresponding to all GVFs for which a cumulant $C=1$ is observed. All GVFs were trained with random agents at levels above (explained below) and, in all, we used approximately $10^7$ actor steps to train this level, a cost that was paid only \emph{once}, as the same model was reused by all agents training the higher levels in specific Android applications.

\begin{figure}
    \centering
    \begin{small}
      \begin{tabular}{l|rrrr}
        \hline
          & Random & DQN & Hierarchy & Human \\
        \hline
        Apple Flinger     & 150 & 0.0$\pm$0.0 &  1899$\pm$276 & 3000 \\
        Blockinger      & 0.0 & 0.0$\pm$0.0 &  44414$\pm$2369 & -  \\
        Catch      & -0.5 & -0.72$\pm$ 0.13 &  0.96$\pm$0.15 & 1 \\
        Classic 2048      & 1000 & 1126$\pm$ 105 &  1615$\pm$161 &  6000 \\
        Dodge      & 0.0 & 0.0$\pm$0.0 &  0.3$\pm$0.09 & 1 \\
        Floodit Easy      & 2 & 0.0$\pm$0.0 &  8.40$\pm$0.28 & 7 \\
        Nostalgic Racer      & 310 & 196$\pm$40 &  372$\pm$41 & 1000 \\
        Vector Pinball      & 9000 & 7478$\pm$1471 &  33240$\pm$8028 & 13000 \\
        \hline
    \end{tabular}
    \end{small}
        \caption{Summary of results at the end of training, compared to human performance and return under a random policy.} \label{fig:empirical:apps:numbers}
\end{figure}

\begin{figure}
    \centering
    \begin{subfigure}[b]{0.35\textwidth}
        \centering
        \includegraphics[width=\textwidth]{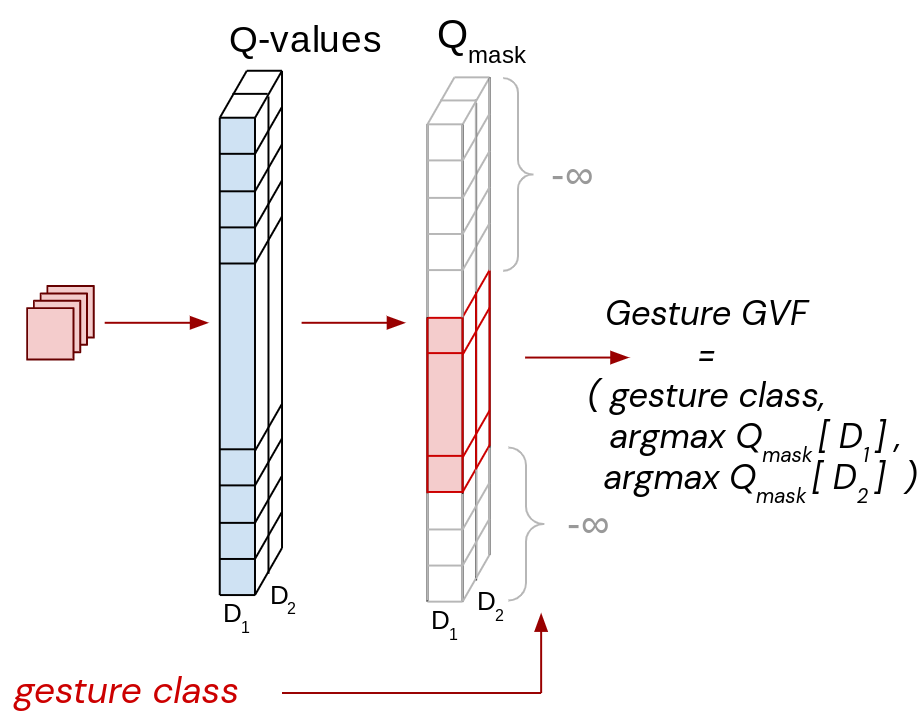}
        \caption{Policy model for GVF Selection Level. The gesture class from the higher level is used to select only a subset of all values. The mask will select slices of size $2 \times 54$, $2 \times 54$, and $2 \times 8$  for the TAP, SWIPE, and FLING classes, respectively. }
        \label{fig:android:selection:policy}
    \end{subfigure}
    \hfill
    \begin{subfigure}[b]{0.3\textwidth}
        \centering
        \includegraphics[width=\textwidth]{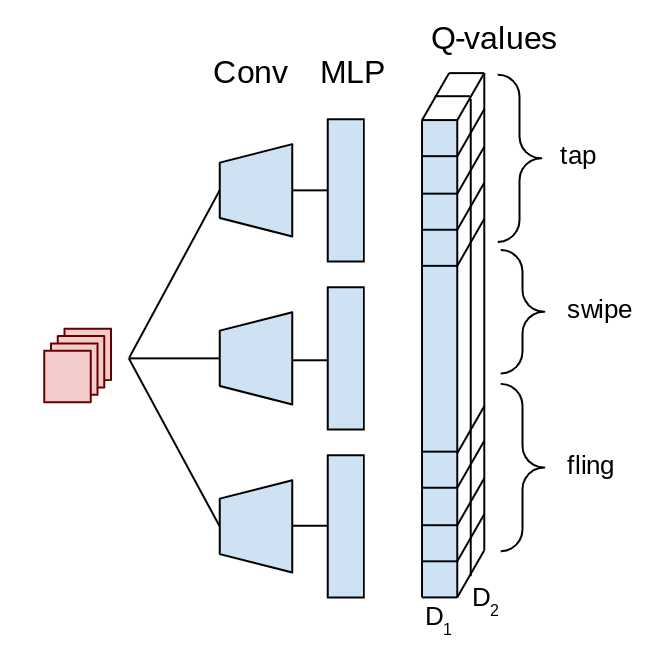}
        \caption{The value model for GVF Selection Level. The input pixel image from AndroidEnv is passed through a convolutional network, followed by a Multi-Layered Perceptron (MLP) to generate a $2 \times 116$ set of Q-values that are used to evaluate each possible GVF choice.}
        \label{fig:android:selection:value}
    \end{subfigure}
    \hfill
    \begin{subfigure}[b]{0.3\textwidth}
        \centering
        \includegraphics[width=\textwidth]{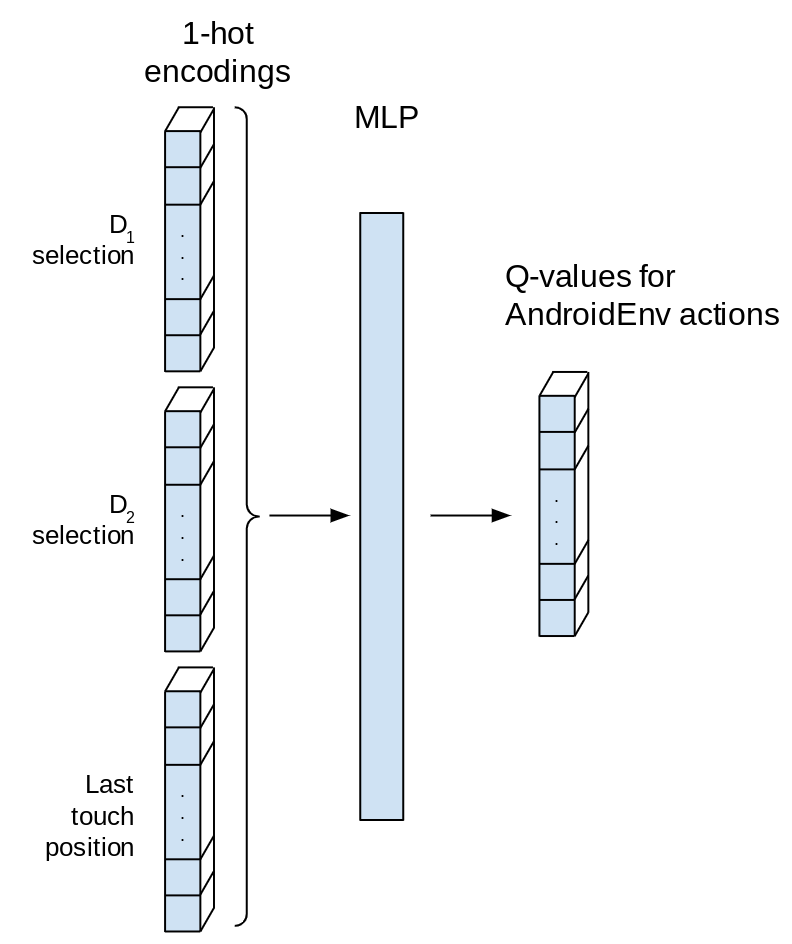}
        \caption{Gesture GVF model. Each of the gesture class is based on a model that takes as input one-hot encodings for the two selections from the higher levels as well as the one-hot encoding of the last touch position. An MLP generates Q-values for each of the $54 \times 2$ AndroidEnv actions.}
        \label{fig:android:gesture:value}
    \end{subfigure}
    \caption{\textbf{Agent models for the AndroidEnv gesture hierarchy.}}
    \label{fig:android:models}
\end{figure}

\paragraph{Level 1: gesture GVF selection.} The second level in the hierarchy uses pixel input data coming from interaction with Android apps to select among all gesture GVFs, which in turn is executed by the lowest level. The level uses the pixel input and reward and the gesture class selection from the upper level to train the corresponding RL agent. The latter combines these signals to generate a parameter, e.g. tap location, for the GVF that should be executed at the lowest level.  
The GVF selection policy is trained using a DQN agent training a joint network for all gesture GVFs. Since the set of swipe GVFs is quite large, i.e. 54 x 54, the Q-value network is designed to output two sets of value estimates: one for the selection of the first parameter out of 54, and another one for the selection of the second parameter. 
See Figures~\ref{fig:android:selection:policy} and~~\ref{fig:android:selection:value} for more details.   

\paragraph{Level 2: gesture class selection.} The third level is trained to select among gesture classes \{tap, swipe, fling\}. The corresponding agent is trained to maximize the average per step reward over the entire episode. This level receives only the environment reward as input and returns one of the three gesture classes. We use the same agent as for the other two layers for training. Since the problem is substantially simpler at this level of abstraction, we used a tabular Q-value representation for the average reward estimations associated with each gesture class.

\section{Discussion}

The results we presented provide strong evidence that task-independent knowledge about the Android action interface, e.g. \emph{finger gestures}, can be used to derive useful hierarchical decompositions. We introduced a flexible and modular signal processing distributed architecture that effectively generates  streams of training data for separate reinforcement learning agents, operating at different levels of abstractions, e.g. selecting a class of GVFs, selecting specific GVFs, executing GVFs. The architecture was used to convert a simple DQN agent into a hierarchy of similar DQN agents, all operating on Android applications, but there is no restriction to this particular choice of agent or environment. Moreover, the hierarchical architecture is not restricted to learning knowledge that is related to finger gestures. In fact, we anticipate even stronger results when the agent is learning abstractions that correspond to more conceptual knowledge on the AndroidEnv platform, e.g. predicting and controlling object movement, menu navigation, affordable interactions with other apps or internet services, discovering common functionalities. Lastly, we believe that the most promising avenue is to allow agents to discover their own collection of GVFs as well as the most appropriate level of abstraction of the knowledge they can capture.

\bibliography{main}

\end{document}